\documentclass[runningheads]{llncs}

\RequirePackage[hyphens]{url} 
\PassOptionsToPackage{hyphens}{url}\usepackage[hidelinks]{hyperref} 

\usepackage[T1]{fontenc}
\usepackage{graphicx}
\usepackage{stackengine}
\usepackage{layouts}

\usepackage{array} 
\usepackage{pbox} 
\usepackage{amsmath} 
\usepackage{amssymb}
\usepackage[super]{nth}
\usepackage{marvosym} 
\usepackage{capt-of}

\usepackage{booktabs}
\usepackage{tabularray}
\usepackage{multirow}

\DeclareMathOperator*{\argmax}{arg\,max}
\DeclareMathOperator*{\argmin}{arg\,min}

\usepackage[table]{xcolor}
\usepackage{pgf}
\usepackage{collcell}

\newcommand*{\MinNumber}{40}%
\newcommand*{\MaxNumber}{200}%

\newcommand{\ApplyGradient}[1]{%
  \pgfmathsetmacro{\PercentColor}{100*(#1-\MinNumber)/(\MaxNumber-\MinNumber)}%
  \edef\x{\noexpand\cellcolor{black!\PercentColor}}\x\textcolor{black}{#1}%
}
\newcolumntype{R}{>{\collectcell\ApplyGradient}{c}<{\endcollectcell}}

\begin{document}

\title{Benchmarking Dependence Measures to Prevent Shortcut Learning in Medical Imaging}

\titlerunning{Preventing Shortcut Learning in Medical Imaging}

\author{
    Sarah M\"uller\inst{1}\textsuperscript{(\Letter)}\and
    Louisa Fay\inst{2,3} \and 
    Lisa M. Koch\inst{1,4} \and
    Sergios Gatidis\inst{2,5} \and
    Thomas K\"ustner\inst{2} \and
    Philipp Berens\inst{1}\textsuperscript{(\Letter)}
}
\authorrunning{S. M\"uller et al.}
\institute{
    Hertie Institute for AI in Brain Health, University of T\"ubingen, Germany \and
    Medical Image and Data Analysis, University Hospital of T\"ubingen, Germany \and
    Institute of Signal Processing and System Theory, University of Stuttgart, Germany \and
    Department of Diabetes, Endocrinology, Nutritional Medicine and Metabolism UDEM, Inselspital, Bern University Hospital, University of Bern, Switzerland \and
    Stanford University, Department of Radiology, Stanford, USA 
    \email{\{sar.mueller,philipp.berens\}@uni-tuebingen.de}
}

\maketitle

\setcounter{footnote}{0}
\begin{abstract}
Medical imaging cohorts are often confounded by factors such as acquisition devices, hospital sites, patient backgrounds, and many more. As a result, deep learning models tend to learn spurious correlations instead of causally related features, limiting their generalizability to new and unseen data. This problem can be addressed by minimizing dependence measures between intermediate representations of task-related and non-task-related variables. These measures include mutual information, distance correlation, and the performance of adversarial classifiers. Here, we benchmark such dependence measures for the task of preventing shortcut learning. We study a simplified setting using Morpho-MNIST and a medical imaging task with CheXpert chest radiographs. Our results provide insights into how to mitigate confounding factors in medical imaging. The project's code is publicly available\footnote{\url{https://github.com/berenslab/dependence-measures-medical-imaging}}.
\end{abstract}

\section{Introduction}
\begin{figure}
    \centering
    \includegraphics{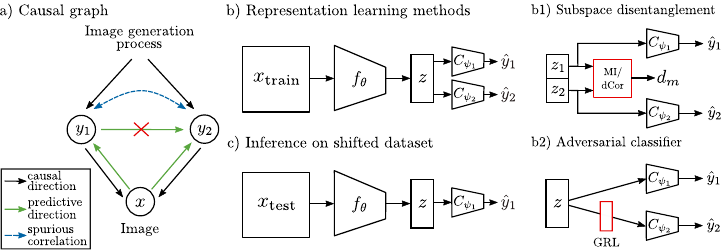}
    \caption{Overview of the causal graph (a) and how the compared methods address the shortcut connection (b). Robust inference performance of task $y_1$ on a shifted data distribution is only possible if the latent space is independent of the confounder $y_2$ (c).}
    \label{fig:overview-figure}
\end{figure}

Medical imaging cohorts are typically heterogeneous and can be confounded by technical factors, including acquisition devices, hospital sites, different patient backgrounds, or study selection bias. These factors may be correlated, and deep learning models run the risk of learning shortcuts based on spurious correlations rather than features causally related to the downstream task \cite{geirhos2020deep,sun2023right} (Fig.\,\ref{fig:overview-figure}a). This means that the models may perform well only within the same data distribution as the training data, but will have reduced performance on a dataset with a different distribution \cite{koch2023subgroup} (Fig.\,\ref{fig:overview-figure}c). For example, the task of disease prediction may be correlated with patient sex for biological reasons or due to selection bias, making it easier in some cases for a deep learning model to infer the disease via image characteristics specific to a patient's sex rather than understanding the hidden causal relationship between the disease and the image features \cite{castro2020causality}.

This problem can be addressed as a representation learning problem with intermediate representations of the data that are independent of known confounding factors. To this end, one can minimize the dependency between the primary task $y_1$ and a spuriously correlated factor $y_2$, which are learned by the classifiers $C_{\psi_1}$ and $C_{\psi_2}$, respectively, from an image encoder latent space $z$ (Fig.\,\ref{fig:overview-figure}b). Subspace disentanglement solves this task by minimizing dependence measures $d_m$ such as mutual information (MI) \cite{belghazi2018mutual} or distance correlation (dCor) \cite{sz2007dCor,zhen2022versatile} between latent subspaces (Fig.\,\ref{fig:overview-figure}b1). Adversarial classifiers \cite{ganin2015unsupervised} (Fig.\,\ref{fig:overview-figure}b2), on the other hand, maximize for primary task $y_1$ performance and at the same time minimize the performance of the correlated factor $y_2$ from a shared latent space, e.g., with a gradient reversal layer (GRL) \cite{ganin2015unsupervised}. Both method categories have been applied to medical imaging problems \cite{fay2023avoid,xie2020mi,hu2024un,mueller2024disentangling,liu2020metrics}, with adversarial classifiers being well-known for domain adaptation tasks \cite{kamnitsas2017unsupervised,he2021adv}. However, existing studies mostly use one individual method, leaving the comparative advantages and disadvantages of different approaches unclear. In our work, we compared different methods regarding their potential to prevent shortcut learning. We first analyzed a toy example using data from Morpho-MNIST \cite{castro2019morpho_mnist}, and later extended our analysis to a medical task using chest radiographs from CheXpert \cite{irvin2019chexpert}.

\section{Methods}\label{sec:methods}
We addressed the problem of shortcut learning by disentangling two highly correlated attributes. We implemented and evaluated three different methods to learn representations invariant to a spuriously correlated factor.

\subsection{Learning Invariant Latent Spaces}
The methods compared in this study differ in how they disentangle the spuriously correlated factor from the task-related variable. However, they can be categorized into two groups: (1) subspace disentanglement, which minimizes a dependence measures between latent subspaces, and (2) adversarial classifiers, which maximize the classification performance of the primary task and minimize the classification performance of the confounding factor on a shared latent space.

\paragraph{Subspace disentanglement} requires a divided latent space into subspaces, where $z_1$ encodes the primary task $y_1$ and $z_2$ the spuriously correlated variable $y_2$. Hence, the objective is to find a mapping $f_\theta(x) = z = (z_1, z_2)$ such that each task can be recovered from the corresponding latent subspaces by a linear mapping $\hat{y}_i = C_{\psi_i} z_i$, but not from the other latent subspaces $z_j$ for $j \neq i$. Optimizing the encoder to map to attribute subspaces is thus equivalent to optimizing the cross-entropy loss for each subspace:
\begin{align}
    (\theta^*, \psi^*) &= \argmin_{\theta, \psi} L_\text{CE,sub}(\theta, \psi), \\
    L_\text{CE,sub}(\theta, \psi) &= \frac{1}{2} \left(y_1^T \log C_{\psi_1} (z_1) + y_2^T \log C_{\psi_2} (z_2) \right).
\end{align}
To make the subspaces statistically independent, the optimization problem is additionally penalized by minimizing a dependence measure $d_m$ between the latent subspaces: 
\begin{align}
    (\theta^*, \psi^*) &= \argmin_{\theta, \psi} L_\text{CE,sub}(\theta, \psi) + \lambda \cdot d_m(z_1, z_2).
\end{align}

Here, we worked with Mutual Information Neural Estimator (MINE) and the empirical distance correlation (dCor) as dependence measure estimates. 
\paragraph{Mutual Information Neural Estimator (MINE)} \cite{belghazi2018mutual} is a lower bound estimator for mutual information. Mutual information (MI) measures the dependence between two variables
\begin{align}
    \text{MI}(z_1, z_2) &= D_\text{KL}(P_{z_1, z_2} \| P_{z_1} P_{z_2})
\end{align}
with $P_{z_1, z_2}$ as the joint probability mass function and $P_{z_1} P_{z_2}$ the product of the marginals. $z_1$ and $z_2$ are independent iff $\text{MI}(z_1, z_2) = 0$. \cite{belghazi2018mutual} lower bounds MI
\begin{align}
    D_\text{KL}(\mathbb{P} \| \mathbb{Q}) \leq \sup_{T\in \mathcal{F}} \mathbb{E}_\mathbb{P}[T] - \log \mathbb{E}_\mathbb{Q}[e^T]
\end{align}
where $\mathcal{F}$ is a family of functions $T_\theta: Z_1 \times Z_2 \rightarrow \mathbb{R}$, parameterized by a deep neural network with parameters $\theta\in\Theta$. Hence, the dependence measure for MINE is defined as 
\begin{align}
    d_{m,\text{MI}}(z_1, z_2) = \sup_{\theta\in\Theta}\mathbb{E}_{P_{z_1, z_2}}[T_\theta] - \log \mathbb{E}_{P_{z_1} P_{z_2}}[e^{T_\theta}].
\end{align}

\paragraph{Distance correlation (dCor)} \cite{sz2007dCor} measures the linear and nonlinear dependence between two random vectors of arbitrary dimension and is bounded in the range $[0, 1]$, where a value of zero means that vectors are independent. In practice, the \textit{empirical distance correlation} can be estimated from batch samples. Consider $N$ samples of subspace vectors $z_1 \in\mathbb{R}^{N\times d_1}$ and $z_2 \in\mathbb{R}^{N\times d_2}$, the distance correlation is defined as
\begin{align}
    d_{m,\text{dCor}}(z_1,z_2) &= \frac{\text{dCov}(z_1, z_2)}{\sqrt{\text{dCov}(z_1, z_1) \, \text{dCov}(z_2, z_2)}},  \text{dCov}(z_1, z_2) = \sqrt{\sum_{i=1}^N\sum_{j=1}^N \frac{A_{i,j} B_{i,j}}{N^2}}
\end{align}
where $A$ and $B$ are the distance matrices. In particular, each element $a_{i,j}$ of the distance matrix $A$ is the Euclidean distance between two samples $\|z_1^{(i)} - z_1^{(j)} \|_2$, after subtracting the mean of row $i$ and column $j$, as well as the matrix mean.

\paragraph{Adversarial (Adv.) Classifiers} address the dependence problem between two variables with a minimax problem: maximizing the primary task performance $y_1$ and minimizing the performance of the spuriously correlated factor $y_2$. In \cite{ganin2015unsupervised}, they employ two classifiers on a shared latent space $z$ (Fig.\,\ref{fig:overview-figure}b2) and enforce invariance of $y_2$ by optimizing
\begin{align}
    L(\theta, \psi_1, \psi_2) &= L_\text{CE}(\theta, \psi_1) - \lambda L_\text{CE}(\theta, \psi_2)\\
    L_\text{CE}(\theta, \psi_i) &= -y_i^T \log C_{\psi_i}(\hat{y_i})
\end{align}
as an adversarial game
\begin{align}
    (\theta^*, \psi_1^*) &= \argmin_{\theta, \psi_1} L(\theta, \psi_1, \psi^*_2) \label{eq:adv_cl_1} \\
    \psi^*_2 &= \argmax_{\psi_2} L(\theta^*, \psi_1^*, \psi_2). \label{eq:adv_cl_2}
\end{align}
In practice, this adversarial game cannot be implemented directly with stochastic gradient descent, but it can be optimized with two different optimizers for Eq.\,\ref{eq:adv_cl_1} and Eq.\,\ref{eq:adv_cl_2} like for generative adversarial networks \cite{goodfellow2014generative}. In \cite{ganin2015unsupervised}, however, they propose a gradient reversal layer (GRL) for optimizing the minimax problem (Fig.\,\ref{fig:overview-figure}b2). The GRL has no parameters, acts as an identity function in the forward pass and inverts the gradient $\frac{\partial L_\text{CE}(\theta, \psi_2)}{\partial \theta}$ during backpropagation.

\subsection{Datasets}
We worked with two datasets: (1) the publicly available $28\times 28$ grayscale images from Morpho-MNIST\footnote{\url{https://github.com/dccastro/Morpho-MNIST}} \cite{castro2019morpho_mnist}, where we selected the thinned and thickened digits of the ``global'' dataset with 39,980 as training and 6,693 test samples. As the primary task $y_1$, we predicted binary digit classes (0-4 versus 5-9) and as the spuriously correlated factor $y_2$ we chose the writing style (thin versus thick) (Tab.\,\ref{tab:experiments}). (2) the publicly available CheXpert\footnote{\url{https://stanfordmlgroup.github.io/competitions/chexpert/}} \cite{irvin2019chexpert} dataset of chest radiographs, which we filtered for frontal images, resulting in 39,979 patients (100,014 images) as training and 827 patients (2,183 images) as test data. Images were first bilinearly resized to a height of 320 pixels and then cropped to a width of 320 pixels from the center. Pleural effusion was the binary primary task and patient's sex was the correlated factor (Tab.\,\ref{tab:experiments}).

\begin{table}[!ht]
    \centering
    \caption{\textbf{Summary of the experiments.} }
    \begin{tblr}{colspec={c|lll}}
        Experiment & Dataset & Primary task $y_1$ & {Correlated factor $y_2$} \\
        \hline
        \# 1 & Morpho-MNIST \cite{castro2019morpho_mnist} & {digits \\ small/high digits} & {writing style \\ thin/thick}\\
        \# 2 & CheXpert \cite{irvin2019chexpert}  & {lung disease \\ healthy/pleural effusion} & {sex \\ female/male} \\
    \end{tblr}
    \label{tab:experiments}
\end{table}

\subsection{Experimental Design}\label{subsec:experimental_design}
\begin{figure}
\stackon{
    \begin{minipage}[b]{0.35\textwidth}
        \centering
        \begin{tabular}[b]{c | c c | c c} 
            & \multicolumn{4}{c}{Experiments} \\
             & \multicolumn{2}{c |}{\# 1} & \multicolumn{2}{c}{\# 2} \\
            \hline
            $y_2$ & \multicolumn{2}{c |}{$y_1$}  & \multicolumn{2}{c}{$y_1$}\\
             & 0 & 1 & 0 & 1 \\
             0 & 9264 & 488 & 10400 & 546\\
             1 & 488 & 9264 & 546 & 10400\\
        \end{tabular}
    \end{minipage}
}
{
    \begin{minipage}[b]{0.35\textwidth}
       \captionof{table}{Absolute label co-occurrences  matrices of training data.}
       \label{tab:co-occurence}
    \end{minipage}
}
\hfill
\stackunder{
    \begin{minipage}[b]{0.55\textwidth}
      \centering
      \includegraphics[width=\textwidth]{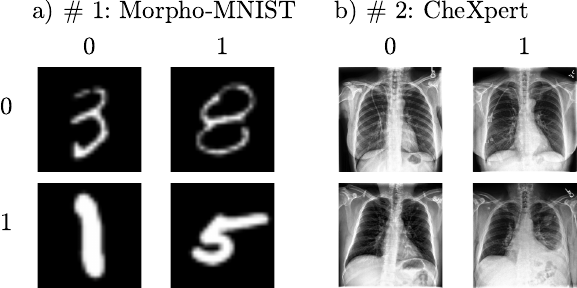}
    \end{minipage}
}
{
    \begin{minipage}[]{0.55\textwidth}
        \captionof{figure}{Sample images from the four label subgroups of the training sets.}
        \label{fig:dataset_examples}
    \end{minipage}
}
\end{figure}

To evaluate each method, we sub-sampled the training data to create strong correlations between the primary task and the spuriously correlated factor \cite{fay2023avoid}. For the training distribution, 95\% of the data was selected from the same category of both labels, resulting in a co-occurrence matrix with 95\% of the data on the main diagonals and 5\% on the off-diagonals (Tab.\,\ref{tab:co-occurence}). For example, for Morpho-MNIST, 95\% of the training images were thin, small digits (label category 0) and thick, high digits (label category 1), with the remaining 5\% from other label combinations. Example images from the four resulting subgroups of Morpho-MNIST and CheXpert are shown in Fig.\,\ref{fig:dataset_examples}a and b, respectively.

We trained all methods with 5-fold cross-validation on the strongly correlated training set. For each fold we selected the best model based on the validation loss and for each method we evaluated the mean performance over the 5-folds on three different data sets:
\begin{itemize}
    \item \textbf{Validation}: Hold-out validation set with the same strong correlations as the training data.
    \item \textbf{Inverted}: Subset of the test data with inverted correlation to the training set, where 95\% of the data are from different label categories.
    \item \textbf{Balanced}: Subset of the test data with no correlation between labels.
\end{itemize}
We expected that methods that prevent shortcut learning would not show a performance drop on the inverted test data compared to the validation data. As a baseline, we compared to an encoder trained with linear classification heads on subspaces, but without any dependence measure minimization (``Baseline''). Additionally, we trained an encoder with classification heads on a balanced version of the correlated training set by oversampling underrepresented subgroups (off-diagonals in the co-occurrence matrix, Tab.\,\ref{tab:co-occurence}) with replacement (``Rebalance'').

\subsection{Implementation Details}
Training was conducted using NVIDIA TITAN Xp and the PyTorch Lightning framework \cite{falcon2019lightning}. A detailed description of all grid searches can be found in the supplementary material. On Morpho-MNIST, we trained the same encoder as in \cite{fay2023avoid} with three convolutional layers and a batch size of 900 samples with an Adam optimizer for 1,000 epochs, mapping to a 4-dimensional latent space (with two 2-dimensional subspaces). For MINE, we trained the encoder for one batch followed by $N_B-1$ batch updates for the MI estimator network, resulting in $N_B \cdot 1,000$ epochs. On CheXpert, we trained a ResNet50 also with a 4-dimensional latent space with batch size 64 for 30 epochs (MINE: $N_B \cdot 30$ epochs) and AdamW. 

\section{Results}
\subsection{Morpho-MNIST}
For each method, we report the mean accuracy over the 5-fold cross-validation on different data distributions (Sec.\,\ref{subsec:experimental_design}). If $z_1$ is not invariant to the spuriously correlated factor, the writing style should still affect digit prediction on shifted distributions. Therefore, we expected methods that prevent shortcut learning to perform better on shifted test distributions, especially for the primary task of digit classification. For the digit classification $y_1$, all methods performed better than the baselines on both the inverted and the balanced distribution (Tab.\ref{tab:results-morpho-mnist-distributions}). MINE had the highest accuracy on the inverted distribution, while the adversarial classifier was best on the balanced distribution. We do not report $z_2$ performance for the adversarial classifier, as it only operates on one shared latent space (Sec.,\ref{sec:methods}). Additionally, we do not report validation performance for rebalancing, since the data was rebalanced before splitting it into folds, making it not comparable with the other methods.

\begin{table}[h!]
    \centering
    \caption{\textbf{Morpho-MNIST: Usage of spurious correlation as shortcut.} Comparison of classification accuracy on different dataset distributions.}
    \begin{tabular}{c | c c c | c c c} 
        & \multicolumn{3}{c|}{$z_1 \rightarrow y_1$ (small/high digits)} & \multicolumn{3}{c}{$z_2 \rightarrow y_2$ (thin/thick)}\\
        Method & Validation & Inverted & Balanced & Validation & Inverted & Balanced \\
        \hline
        Baseline & 98.2 & 79.3 & 86.4 & \textbf{99.9} & 98.8 & 99.5 \\
        Rebalance & - & 88.4 & 91.5 & - & 99.1 & 99.6 \\
        MINE & 93.5 & \textbf{94.2} & 91.6 & 99.7 & \textbf{99.6} & \textbf{99.7}\\
        dCor & \textbf{98.4} & 87.1 & 91.2 & \textbf{99.9} & 99.0 & 99.5\\
        Adv. Classifier & 97.6 & 89.7 & \textbf{92.5} & - & - & -\\
    \end{tabular}
    \label{tab:results-morpho-mnist-distributions}
\end{table}

\begin{figure}[h]
    \centering
    \includegraphics[width=\textwidth]{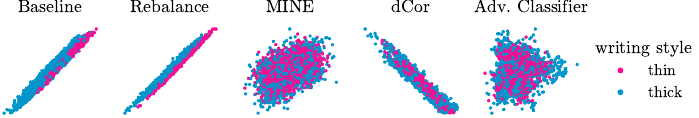}
    \caption{Latent subspace $z_1$ trained on Morpho-MNIST, colored by writing style $y_2$.}
    \label{fig:morpho-mnist-embeddings}
\end{figure}

We evaluated the disentanglement performance based on the the predictive subspace performance using a kNN classifier, resulting in a confusion matrix of mean accuracies for each subspace-label-combination (Tab.\,\ref{tab:results-morpho-mnist-encoding-knn}). Since we evaluated the classifiers on the balanced test data, a method with good disentanglement performance should have accuracies on the off-diagonals that are close to random guessing (50\%). Although data rebalancing showed good performance for the shifted test distributions (Tab.\,\ref{tab:results-morpho-mnist-distributions}), the disentanglement performance was comparable to the baseline method, since one of the off-diagonals is 20.6\% from random guessing. MINE achieved the best overall performance, followed by the adversarial classifier and dCor. These findings were supported by a qualitative evaluation of the writing style encoding in the latent space $z_1$ (Fig.\,\ref{fig:morpho-mnist-embeddings}).

\begin{table}[t]
    \centering
    \caption{\textbf{MorphoMNIST: Confusion matrix for disentanglement performance.} Confusion matrix of kNN accuracy (k=30) on the \textit{balanced} dataset.}
    \begin{tabular}{ c c | R  R | R R | R  R | R R | R c| }
          Method & & 
          \multicolumn{2}{c|} {Baseline} &
          \multicolumn{2}{c|} {Rebalance} &
          \multicolumn{2}{c|} {MINE} &
          \multicolumn{2}{c|} {dCor} &
          \multicolumn{2}{c|} {Adv. Classifier} \\
          \hline
          \begin{tabular}{@{}c@{}} Subspaces / \\ Labels \end{tabular} & &
          \multicolumn{1}{c} {$z_1$} &
          \multicolumn{1}{c|} {$z_2$} &
          \multicolumn{1}{c} {$z_1$} &
          \multicolumn{1}{c|} {$z_2$} &
          \multicolumn{1}{c} {$z_1$} &
          \multicolumn{1}{c|} {$z_2$} &
          \multicolumn{1}{c} {$z_1$} &
          \multicolumn{1}{c|} {$z_2$} &
          \multicolumn{1}{c} {$z_1$} &
          \multicolumn{1}{c|} {$z_2$} \\
          $y_1$ &  & 86.5 &  50.5 & 90.9 & 50.2 & 91.4 & 50.1 &  91.0 & 52.3 & 92.4 & -\\
          $y_2$ &  & 69.8 &  99.5 & 70.6 & 99.5 & 48.9 & 99.7 &  57.8 & 99.5 & 55.3 & - \\
    \end{tabular}
\label{tab:results-morpho-mnist-encoding-knn}
\end{table}

\subsection{CheXpert}
For our CheXpert experiments, we report mean area under the receiver operating characteristic (AUROC) over the 5-fold cross-validation of each method. We expected methods that prevent shortcut learning to perform better on inverted and balanced test distributions for the disease classification. Here, we saw that for pleural effusion classification all methods except dCor outperformed the baseline on the shifted test distributions, with MINE performing best (Tab.\ref{tab:results-chexpert-distributions}). Compared to the experiments with Morpho-MNIST, both dCor and the adversarial classifier showed a drop in performance, with dCor only improving in balanced performance.

\begin{table}[h!]
    \centering
    \caption{\textbf{CheXpert: Usage of spurious correlation as shortcut.} Comparison of classification AUROC on different dataset distributions}
    \begin{tabular}{c | c c c | c c c} 
        & \multicolumn{3}{c|}{$z_1 \rightarrow y_1$ (healthy/disease)} & \multicolumn{3}{c}{$z_2 \rightarrow y_2$ (female/male)}\\
        Method & Validation & Inverted & Balanced & Validation & Inverted & Balanced \\
        \hline
        Baseline & \textbf{98.5} & 45.8 & 80.5 & \textbf{99.5} & 96.8 & \textbf{97.9} \\
        Rebalance & - & 79.8 & 90.0 & - & 96.2 & 96.8  \\
        MINE & 93.8 & \textbf{87.7} & \textbf{92.4} & 98.5 & \textbf{97.6} & 96.4  \\
        dCor & 98.3 & 53.4 & 83.0 & \textbf{99.5} & 94.7 & 97.2 \\
        Adv. Classifier & 95.5 & 67.4 & 84.9 &  - & - & - \\
    \end{tabular}
    \label{tab:results-chexpert-distributions}
\end{table}

\begin{figure}
    \centering
    \includegraphics[width=\textwidth]{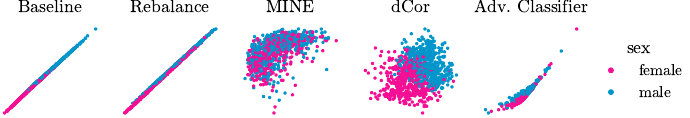}
    \caption{Latent subspace $z_1$ trained on CheXpert, colored by sex $y_2$.}
    \label{fig:chexpert-embeddings}
\end{figure}

To evaluate subspace disentanglement with CheXpert, we again computed the confusion matrix of kNN mean accuracies for all subspace-label-combinations. Here, data rebalancing outperformed the baseline, with off-diagonals dropping by 2.5\% and 12.5\%. MINE had the best overall performance, followed by rebalancing and the adversarial classifier. dCor showed no disentanglement effect, with off-diagonals comparable to the baseline, which was also visible in the qualitative evaluation of the disease encodings in Fig.\,\ref{fig:chexpert-embeddings}.

\begin{table}[h!]
    \centering
    \caption{\textbf{CheXpert: Confusion matrix for disentanglement performance.} Confusion matrix of kNN accuracy (k=30) on the \textit{balanced} dataset.}
    \begin{tabular}{ c c | R  R | R R | R  R | R R | R c | }
          Method & & 
          \multicolumn{2}{c|} {Baseline} &
          \multicolumn{2}{c|} {Rebalance} &
          \multicolumn{2}{c|} {MINE} &
          \multicolumn{2}{c|} {dCor} &
          \multicolumn{2}{c|} {Adv. Classifier} \\
          \hline
          \begin{tabular}{@{}c@{}} Subspaces / \\ Labels \end{tabular} & &
          \multicolumn{1}{c} {$z_1$} &
          \multicolumn{1}{c|} {$z_2$} &
          \multicolumn{1}{c} {$z_1$} &
          \multicolumn{1}{c|} {$z_2$} &
          \multicolumn{1}{c} {$z_1$} &
          \multicolumn{1}{c|} {$z_2$} &
          \multicolumn{1}{c} {$z_1$} &
          \multicolumn{1}{c|} {$z_2$} &
          \multicolumn{1}{c} {$z_1$} &
          \multicolumn{1}{c|} {$z_2$} \\
          $y_1$ &  & 71.8 &  56.3 & 81.4 & 53.8 & 85.4 & 48.0 &  73.7 & 59.8 & 78.5 & -\\
          $y_2$ &  & 76.5 &  92.8 & 64.0 & 89.9 & 56.4 & 92.4 &  75.7 & 91.5 & 67.6 & -\\
    \end{tabular}
\label{tab:results-chexpert-encoding-knn}
\end{table}

\section{Discussion}
In our quantitative benchmark, MINE was the best method at preventing shortcut learning and improving disentanglement between the primary task and a spuriously correlated variable. However, MINE's training duration is longer due to the asynchronous training of the MI estimator network and the encoder. Dataset rebalancing was surprisingly effective, especially for primary task performance on the inverted test distribution, but its disentanglement performance was sub-optimal for Morpho-MNIST. Both dCor and the adversarial classifier, although fast to evaluate, showed lower performance on the medical task, indicating limited robustness across domains. A key limitation for all methods is the need for supervision to encode attributes into latent subspaces. In future work we want to study additional medical datasets, explore the role of correlation strength, and evaluate other measures such as Maximum Mean Discrepancy \cite{gretton2012mmd}.

\begin{credits}
\subsubsection{\ackname} 
This project was supported by the Hertie Foundation and by the Deutsche Forschungsgemeinschaft under Germany’s Excellence Strategy with the Excellence Cluster 2064 ``Machine Learning — New Perspectives for Science'', project number 390727645. This research utilized compute resources at the T\"ubingen Machine Learning Cloud, INST 37/1057-1 FUGG. PB is a member of the Else Kr\"oner Medical Scientist Kolleg ``ClinbrAIn: Artificial Intelligence for Clinical Brain Research''. The authors thank the International Max Planck Research School for Intelligent Systems (IMPRS-IS) for supporting SM.
\end{credits}

\bibliographystyle{splncs04_without_urls}
\bibliography{references.bib}

\newpage

\section*{Appendix}

\subsection*{Hyperparameters and Grid Searches}
\paragraph{Morpho-MNIST:} For MINE, we took the hyperparameters $N_B=3$ and $\lambda=0.55$ from \cite{fay2023avoid}. For dCor, we performed a hyperparameter grid search for $\lambda \in \{0.1 n | n\in\mathcal{Z}, 1 \leq n \leq 10\}$ (best performance for $\lambda=0.5$). For the adversarial classifier, we used all the hyperparameters described in \cite{ganin2015unsupervised}, including SGD optimizer with $0.9$ momentum, a start learning rate of $0.01$, a learning rate annealing, and a schedule for 
\begin{align*}
    \lambda = \alpha \cdot \left(\frac{2}{1+\exp(-\gamma\cdot p)}  - 1\right)
\end{align*}
where $p$ is the training progress linearly changing from 0 to 1.
For the rest of the methods, we used the Adam optimizer with a learning rate of $0.001$, except for MINE, where the learning rate was $0.0001$ for both the encoder and the MI estimator network.

\paragraph{CheXpert:} We trained all methods with an AdamW optimizer with a learning rate of 0.001, except for the adversarial classifier where used the SGD optimizer with parameters as in \cite{ganin2015unsupervised} and a start learning rate of 0.002. For MINE, we performed a grid search for $N_B\in [3,5]$ and $\lambda \in [0.3, 0.5]$ (best model for $N_B=5, \lambda=0.5$) and set the epochs to $N_B \cdot 30$. For the adversarial classifier, we searched in $\alpha \in\{0.2 n | 1 \leq n \leq 5\}$ and $\gamma \in\{n | 4 \leq n \leq 7\}$ (best model for $\alpha=0.4$, $\gamma=4$) and for dCor we performed the same grid search as for the Morpho-MNIST experiments (best model for $\lambda=0.1$).

\end{document}